\documentclass[runningheads]{llncs}

\usepackage{eccv}

\usepackage{eccvabbrv}
\usepackage{colortbl}
\usepackage{graphicx}
\usepackage{booktabs}
\usepackage{adjustbox}
\usepackage{multirow}
\usepackage[accsupp]{axessibility}  % Improves PDF readability for those with disabilities.

\usepackage{hyperref}

\usepackage{orcidlink}

\begin{document}

\title{ScatterFormer: Efficient Voxel Transformer with Scattered Linear Attention} 

\titlerunning{ScatterFormer}

% Second Author\inst{2,3}\orcidlink{1111-2222-3333-4444} \and
% Third Author\inst{3}\orcidlink{2222--3333-4444-5555}}
\author{
Chenhang He\inst{1} \and 
Ruihuang Li\inst{1,2} \and
Guowen Zhang\inst{1} \and
Lei Zhang\inst{1,2}}
\authorrunning{C. He et al.}

\institute{
The Hong Kong Polytechnic University, Hong Kong SAR, China  \and
OPPO Research, Shenzhen, China \\
\email{chenhang.he@polyu.edu.hk} \\
\email{csrhli@comp.polyu.edu.hk} \\
\email{guowen.zhang@connect.polyu.edu.hk} \\
\email{cslzhang@comp.polyu.edu.hk}
}

\maketitle

\begin{abstract}
Window-based transformers excel in large-scale point cloud understanding by capturing context-aware representations with affordable attention computation in a more localized manner. However, the sparse nature
of point clouds leads to a significant variance in the number of voxels per window. Existing methods group the voxels in each window into fixed-length sequences through extensive sorting and padding operations, resulting in a non-negligible computational and memory overhead. In this paper, we introduce ScatterFormer, which to the best of our knowledge, is the first to directly apply attention to voxels across different windows as a single sequence. The key of ScatterFormer is a Scattered Linear Attention (SLA) module, which leverages the pre-computation of key-value pairs in linear attention to enable parallel computation on the variable-length voxel sequences divided by windows. Leveraging the hierarchical structure of GPUs and shared memory, we propose a chunk-wise algorithm that reduces the SLA module's latency to less than 1 millisecond on moderate GPUs.  Furthermore, we develop a cross-window interaction module that improves the locality and connectivity of voxel features across different windows, eliminating the need for extensive window shifting.  Our proposed ScatterFormer demonstrates 73.8 mAP (L2) on the Waymo Open Dataset and 72.4 NDS on the NuScenes dataset, running at an outstanding detection rate of 23 FPS. The code is available at \href{https://github.com/skyhehe123/ScatterFormer}{\color{red}{https://github.com/skyhehe123/ScatterFormer}}.

\end{abstract}

\section{Introduction}
\label{sec:intro}

\begin{figure}[t]
    \centering
    \begin{subfigure}{0.7\columnwidth}
        \centering
        \includegraphics[width=0.99\linewidth]{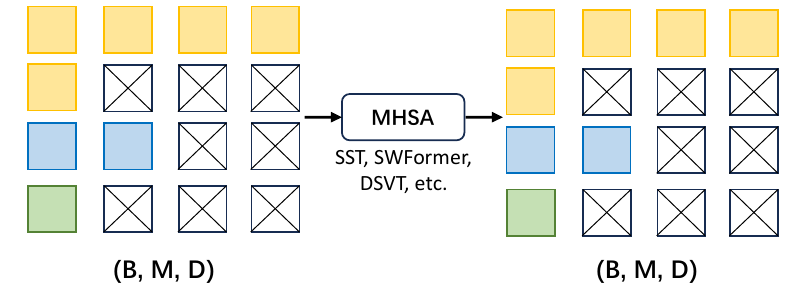}
        \vspace{-7mm}
        \caption{}
        \vspace{3mm}
        \label{fig:traditional}
    \end{subfigure}
    
    \begin{subfigure}{0.72\columnwidth}
        \centering
        \includegraphics[width=0.99\linewidth]{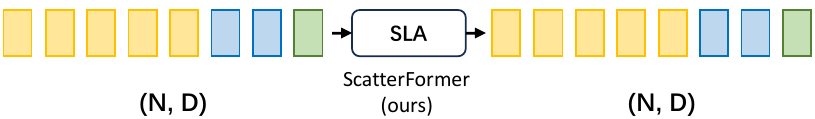}
        \vspace{-4mm}
        \caption{}
        \label{fig:linear}
    \end{subfigure}
    \vspace{-2mm}
    \caption{Illustration of (a) group-based attention and (b) our scattered linear attention on variable-length sequences. The square in different colors represent voxels in different windows and the square with a cross represents the padding voxel.}
    \vspace{-6mm}
    \label{fig:motivate}
\end{figure}

In the field of 3D object detection, the use of point clouds has become increasingly popular, especially for providing accurate and reliable perception results in autonomous systems. Unlike image data, point clouds obtained from LiDAR are often sparse and nonuniformly distributed, with varying density depending on their distances from the sensor. Earlier approaches utilize point-cloud operators like PointNet++ \cite{pointnet++} for feature extraction in continuous space \cite{pointrcnn}. Some approaches \cite{second, voxelnet, centerpoint, pointpillars, pixor, sa-ssd, fsd, voxelnext} have transformed point clouds into voxel grids, which are then efficiently processed by a sparse convolutional neural network (SpCNN) \cite{second}.

Recently, the success of vision transformers \cite{vit, swin} has motivated a number of attention-based methods to process indoor point clouds \cite{pct, pointformer, pointTr, 3detr, votr, group-free}. Inspired by SwinTransformer \cite{swin}, various studies \cite{sst, flatformer, voxset, dsvt} have advanced the application of window-based Transformers in large-scale 3D detection tasks within outdoor environments, achieving outstanding performance and highlighting their potential as alternatives to SpCNN. However, due to the inherent sparsity of point clouds, the number of features grouped by windows can vary significantly, hampering the parallelism in attention computation. To resolve this issue, SST \cite{sst} and SWFormer \cite{swformer} group the voxels within a window into different batches and manages the attention computation in a hybrid serial-parallel manner. More effectively, DSVT \cite{dsvt} alternatively sort the voxels within a window from different axes and partition them into sequences of fixed length, which allow parallel computation on sparse voxels. Albeit effective, these group-based methods, as shown in Figure \ref{fig:motivate}\textcolor{red}{(a)}, require extensive sorting and padding operations, incurring substantial memory and computational overhead. 

In this paper, we delve into the window-based voxel transformer where the voxels grouped by windows form variable-length sequences $\{X_1 \in \mathbb{R}^{n_1\times d}, X_2 \in \mathbb{R}^{n_2\times d}, ..., X_k \in \mathbb{R}^{n_k\times d}\}$. Their corresponding self-attention matrices $\{A_1 \in \mathbb{R}^{n_1\times n_1}$, $A_2 \in \mathbb{R}^{n_2\times n_2}$,..., $A_k \in \mathbb{R}^{n_k\times n_k}\}$ occupy irregular memory spaces, making it a challenge to perform attention in parallel for the voxels of the entire scene. Recent efforts on linear attention \cite{xcit, flattenTr, linformer, transformer-rnn} have emerged a promising alternative to traditional attention. By approximating
softmax operation with the kernel function, \ie, $\phi(Q) \cdot \phi(K)^\mathsf{T}  \approx  \text{softmax} (QK^\mathsf{T})$, we can change the computation order from $(Q \cdot K^\mathsf{T})\cdot V$ to $Q \cdot (K^\mathsf{T}\cdot V)$. This not only results in linear complexity but also yields a compressed hidden state $S \in \mathbb{R}^{d\times d}$ that is invariant to the sequence length. Another attractive property of linear attention is that it can be converted into a ``recurrent'' form:
\begin{equation}
   o_t = q_tS_t; \quad S_t = S_{t-1} + k_t^{\mathsf{T}}v_t
   \label{eq:recurrent}
\end{equation}
where the sequences can be divided into non-overlapping chunks, $\{q_j, k_j, v_j\}_{j=1:t}$. The output can be calculated by scanning over the sequence, based on an updated hidden state $S_t$. This enables the use of chunk-level computation to address the inability to parallelize at the sequence level due to varying lengths.

Inspired by this, we introduce the \textbf{Scattered Linear Attention} (SLA) module, which accommodates linear attention in a window-based voxel transformer. As shown in Figure \ref{fig:motivate}\textcolor{red}{(b)}, the SLA module treats the voxels of the entire scene into a single sequence and processes them directly without padding voxels. Using the recurrent form of linear attention, we develop an I/O-aware algorithm to further optimize matrix multiplication on voxel sequences. Specifically, we divide the voxel sequence into multiple chunks and loaded them into the shared memory (SRAM) of the GPU. The computation of the hidden state matrix in each window is then achieved through a series of chunk-wise matrix multiplications and cumulative sums. This optimized implementation significantly reduces I/O overhead and memory usage, resulting in extremely fast and memory-efficient attention computation.

In addition, current window-based transformer models use window shifting to propagate the information across the windows. After performing an instruction-level analysis of existing implementations, we observe that the voxel permutation operations caused by window shifting consume significant computational overhead. To address this issue, we propose a cross-window interaction (CWI) module, which is composed of depth-wise convolutions with small 2D kernels and lengthy 1D kernels that allow the voxel features in each window to fully interact with the features in other windows. As a result, the proposed CWI module can improve both the locality and connectivity of the voxel features while requiring minimal computational effort. 

Building upon the SLA and CWI modules, we propose the \textbf{ScatterFormer}, an innovative voxel transformer for large-scale point cloud understanding. Our experiments show that ScatterFormer achieves linear complexity without compromising accuracy. It achieves superior results compared to the state-of-the-art model DSVT \cite{dsvt}. The latency of ScatterFormer is significantly lower than that of transformer-based detectors \cite{sst, dsvt, voxset}, which is comparable to that of sparse convolution-based detectors \cite{centerpoint}.

In conclusion, we delve into attention on voxels grouped by windows, highlighting the challenges in memory allocation and computation for variable-length sequences. Then, we introduce an SLA module, which can directly process the voxels grouped by windows by facilitating the linear attention formula. A chunk-wise matrix multiplication algorithm is proposed to further accelerate the attention computation in SLA. Finally, we present ScatterFormer, which has linear complexity and can efficiently process large-scale LiDAR scenes, achieving better accuracy with lower latency compared to existing transformer-based detectors.

\section{Related Work}
\label{sec:related}

%-------------------------------------------------------------------------
\subsection{Point Cloud-based 3D Object Detection}

There exist two primary point-cloud representations in 3D object detection, \ie, point-based and voxel-based ones. In point-based methods \cite{pointrcnn, pointgnn, pcrgnn, frustum, votenet}, point clouds are first passed
through a point-based backbone network \cite{pointnet}, in which the points
are gradually sampled and features are learned by point cloud
operators. F-Pointnet \cite{frustum} first employs PoinNet to detect 3D objects based on the frustums lifted by 2D proposals. PointRCNN \cite{pointrcnn} direct generates 3D proposals from point-based features with foreground segmentation. VoteNet \cite{votenet} clusters objects from the surface points using a deep Hough voting method. Other point operators based on the point graph \cite{pointgnn, pcrgnn} and the range view \cite{rangedet} have also been developed for point cloud processing. 

Point-based methods are primarily limited by their inference efficiency and the absence of contextual features in a continuous space. On the other hand, voxel-based approaches \cite{second, voxelnet, voxelnext, pointpillars, centerpoint, sa-ssd, se-ssd} transform the entire point clouds into regular grids through voxelization, showcasing superior efficiency and context representation. VoxelNet \cite{voxelnet} proposes a voxel feature encoding (VFE) module and combines it with 3D convolutions to extract voxel features in an end-to-end manner. SECOND \cite{second} optimizes 3D convolution for sparse data, resulting in a significant reduction in both time and memory. PointPillars \cite{pointpillars} demonstrates an efficient model by stacking voxels into vertical columns, subsequently processing them with naive 2D convolutions. There are also efforts \cite{sa-ssd, pvrcnn} that explore hybrid representations of point-based and voxel-based networks, demonstrating a better trade-off between speed and accuracy. However, a common limitation across these methods is their reliance on small convolution kernels with restricted receptive fields, making them less adept at capturing the global context for 3D object detection.

In this paper, we shed light on the voxel-based methods and introduce a Scattered Linear Attention module that enables dynamic modeling among voxels within each window with linear complexity, thereby efficiently extracting window-based contextual features.

\subsection{Transformer on Point Cloud}
Inspired by the significant success of self-attention in NLP \cite{transformer} and CV \cite{vit, deit}, Transformers have been adapted for 3D vision due to their ability to capture long-range dependencies. The Point Transformers \cite{pointTr, pointformer} employ attention to modulate point clouds for classification and segmentation tasks. PCT \cite{pct} presents an optimized offset attention module, which, when combined with the implicit Laplacian operator and normal estimation, becomes more adept at point-cloud processing. Some methods like \cite{sst, group-free} opt for voxels or key points to optimize latency. 3DETR \cite{3detr} proposes a query-based 3D object detection scheme. CT3D \cite{ct3d} enhances the region-based network using a channel-wise Transformer framework. To achieve context-rich representations, several studies \cite{votr, dsvt, flatformer, sst, voxset} have integrated the attention module into point- or voxel-based encoders. For instance, VoTr \cite{votr} utilizes dilated attention for expanded receptive fields; VoxSet \cite{voxset} applies set attention for extracting point-based features in set-to-set translation; SST \cite{sst} employs local attention with shifted windows; and OcTr \cite{octr} adopts an Octree-based attention for efficient hierarchical context learning. Nevertheless, how to efficiently leverage global context from attention remains a challenge due to the intrinsic sparsity of point clouds. DSVT \cite{dsvt} and FlatFormer \cite{flatformer} group the voxels within each window into a series of fixed-length voxel sets, thus extracting the features in a fully parallel manner. Recently, a group-free state-space model \cite{zhang2024voxel} was proposed to directly process voxels as a sequence. However, these approaches more or less lose the spatial proximity and incur extensive computational overhead in grouping and sorting the voxels. 

\section{Method}
\label{sec:method}
\begin{figure}[t]
    \centering
    \includegraphics[width=0.99\columnwidth]{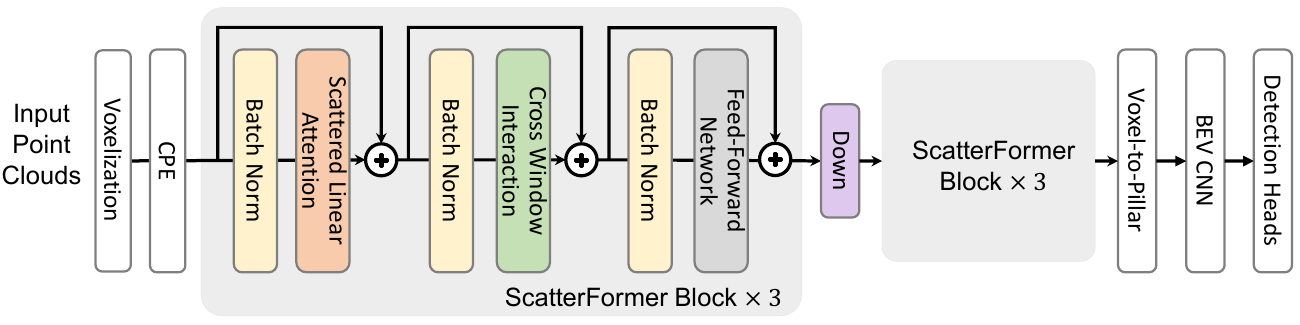}
    \caption{The macro design of ScatterFormer. The backbone comprises a Conditional Position Encoding (CPE) and six transformer blocks. Each block is composed by a Scattered Linear Attention (SLA) module, a Cross-Window Interaction (CWI) module, and a Feed-Forward Network (FFN).}
    \vspace{-5mm}
    \label{fig:macro}
\end{figure}
The overall architecture of our proposed ScatterFormer is depicted in Figure \ref{fig:macro}. It begins with the input point clouds, which are voxelized and transformed into high-dimensional embeddings using a VFE layer \cite{voxelnet}. These embeddings are then processed through Conditional Positional Encoding (CPE) using a shallow convolutional network \cite{cpvt}. The encoded features enter the ScatterFormer backbone, consisting of six ScatterFormer blocks. Each block includes a Scattered Linear Attention (SLA) module, a Cross-Window Interaction (CWI) module, and a Feed-Forward Network (FFN), interspersed with Batch Normalization layers and skip connections. After three ScatterFormer blocks, the voxel features are downsampled via a sparse convolutional layer. The downsampled features are then converted into pillar features \cite{dsvt}, generating compact BEV features for bounding-box prediction. ScatterFormer ScatterFormer stands out by not requiring voxel features to be organized into fixed-length sets \cite{dsvt, sst, flatformer}, enabling flexible attention computation across windows. Additionally, the CWI module obviates the need for window shifting. These innovations allow ScatterFormer to avoid unnecessary memory allocation and permutation operations, achieving high efficiency comparable to CNN-based models.

\subsection{Linear Attention}
We begin by revisiting the self-attention introduced by \cite{transformer}. Given an input matrix $x \in \mathbb{R}^{N \times d}$,  where $N$ denotes the number of tokens and $d$ denotes the dimensionality. Self-attention first linearly projects the input to queries, keys, and values using weight matrices $W_Q$, $W_K$, and $W_V$, such that $ Q = xW_Q, K = xW_K, V = xW_V$. Then it computes a Softmax-based attention map based on queries and keys \ie, $A(Q, K) = \text{Softmax}(QK^\mathsf{T})/\sqrt{d}$. The output of self-attention is defined as the weighted sum of $N$ values with the weights corresponding to the attention map, \ie, $O = A(Q, K)V$. This approach involves calculating the similarity for all query-key pairs, which yields a computational complexity of $O(N^2)$. 

Linear attention \cite{transformer-rnn, xcit} offers a notable alternative to self-attention by significantly reducing its computational complexity from $O(N^2)$ to $O(N)$. This efficiency is achieved through the application of kernel functions on the query ($Q$) and key ($K$) matrices, which effectively approximates the original attention map without relying on Softmax, \ie, $A(Q, K) = \phi(Q) \phi(K)^\mathsf{T}$. Taking advantage of the associative property of matrix multiplication, the computation shifts from $(\phi(Q) \phi(K)^\mathsf{T}) V$ to $\phi(Q) (\phi(K)^\mathsf{T} V)$. This modification leads to a computational complexity proportional to the number of tokens, maintaining an order of $O(N)$.

\subsection{Scattered Linear Attention (SLA)}
\begin{figure*}[t!]
     \centering
    \includegraphics[width=0.99\textwidth]{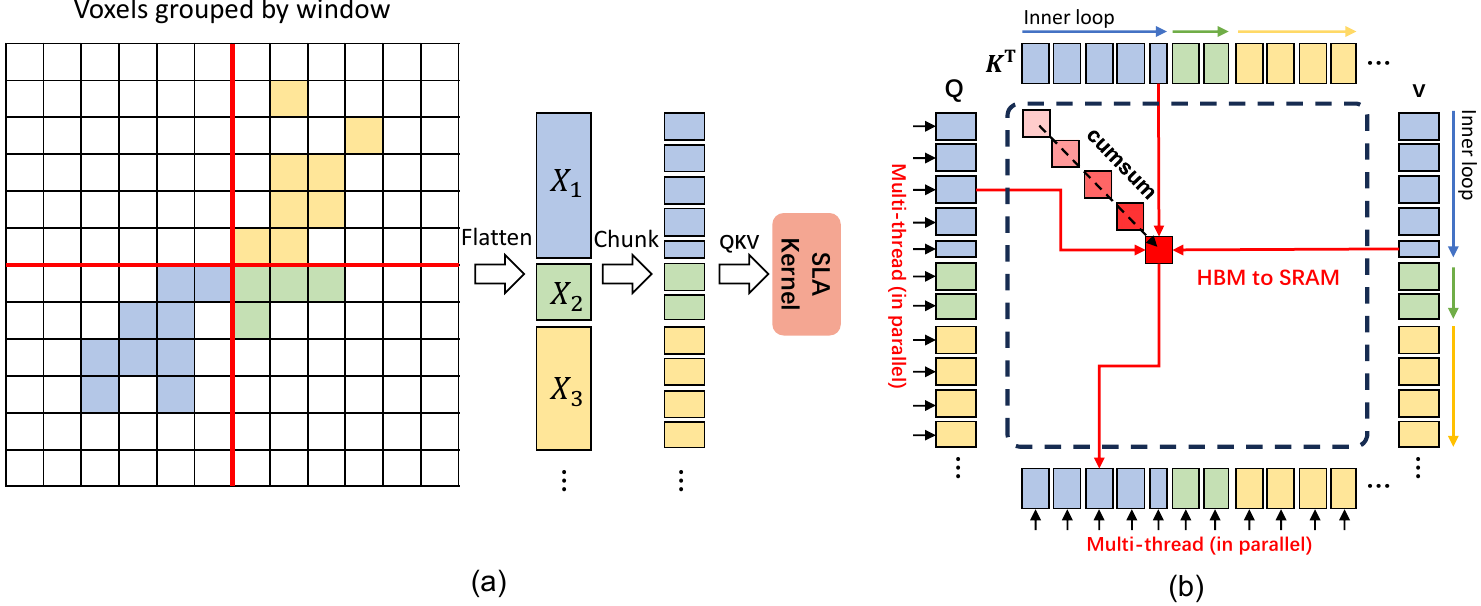}
    \vspace{-1mm}
     \caption{(a) The voxel features are sorted by windows and flattened to a single matrix, which is further partitioned into multiple chunks of uniform length and sent to SLA kernel for attention computation. (b) In SLA, we first allocate individual threads to each window, each of them iterates over all corresponding key and value chunks, calculates, and accumulates their products to obtain the hidden state matrix of each window. Then, multiple threads are allocated to these query chunks $q_i \in Q$, calculating the chunk-wise output by multiplying $q_i$ with the corresponding hidden state matrix. }
    \label{fig:main}
    \vspace{-3mm}
\end{figure*}

Instead of organizing the voxels within a window into fixed-length subsets, we arrange the voxels of the entire scene into a single flattened matrix, as illustrated in Figure \ref{fig:main}\textcolor{red}{(a)}. These voxels are ordered based on their window coordinates, ensuring that voxels from the same window form a sub-matrix in contiguous memory. Initially, we use a shared projection layer to map all the voxels in the scene to query, key, and value representations, denoted $Q$, $K$, $V$. Subsequently, we perform attention computations on these submatrices separately, resulting in the scattered linear attention (SLA) formula as follows:
\begin{equation}
\text{SLA}(Q, K, V) =\text{Concat} [\text{LA}(Q^{j}, K^{j}, V^{j})]_{ j=1:M},
\label{eq:scatter}
\end{equation}
where $M$ is the number of non-empty window in the current scene and $\text{LA}$ is the linear attention formula used in \cite{transformer-rnn}. Based on this formula, the output voxel features $O^j$ in the $j^{\text{th}}$ window can be defined as:
\begin{equation}
\text{LA}: O^j = \frac{\phi(Q) \sum_{i=1}^{{m^j}} \phi(K_i)^{\top} V_i}{\phi(Q) \sum_{i=1}^{{m^j}} \phi(K_i)^{\top}}
\label{eq:la}
\end{equation}
where $m^j$ is the number of voxel in the window and $\phi(x)$ is the kernel function.

\vspace{3mm}
\noindent \textbf{Hardware Efficient Implementation.}
It should be noted that the implementation of Equation \ref{eq:scatter} is not straightforward, as the sub-matrices have different numbers of rows. One naive implementation is to cache the ``kv'' matrix of each voxel and apply \textit{scatter}\footnote{https://github.com/rusty1s/pytorch\_scatter} operator to accumulate them within each window. However, this would consume a high memory overhead and IO latency. Considering that GPUs feature a memory hierarchy that includes larger, slower global GPU memory (high-bandwidth memory; HBM) and smaller, faster shared memory (SRAM), optimal utilization of SRAM to minimize HBM I/O costs can lead to significant speed-ups. Based on this, we partition the flattened $Q,K,V$ into multiple chunks, load them from slow HBM to fast SRAM. Specifically, for each window, we allocate an single thread. Each thread iterates over all the key and value chunks, calculates, and accumulates their products to obtain the hidden state matrix of the current window. After calculating the hidden state matrix in window, we assign individual thread to each query chunk $q_i \in Q^j$ and calculate the chunk-wise output can then be following Equation \ref{eq:la}. Figure \ref{fig:main}\textcolor{red}{(b)} illustrates this chunk-wise matrix multiplication approach. As can be seen, it not only avoids outputting large matrices but also allows for more flexible processing of variable-length sequences.

\begin{figure}[t]
    \centering
    \includegraphics[width=0.8\columnwidth]{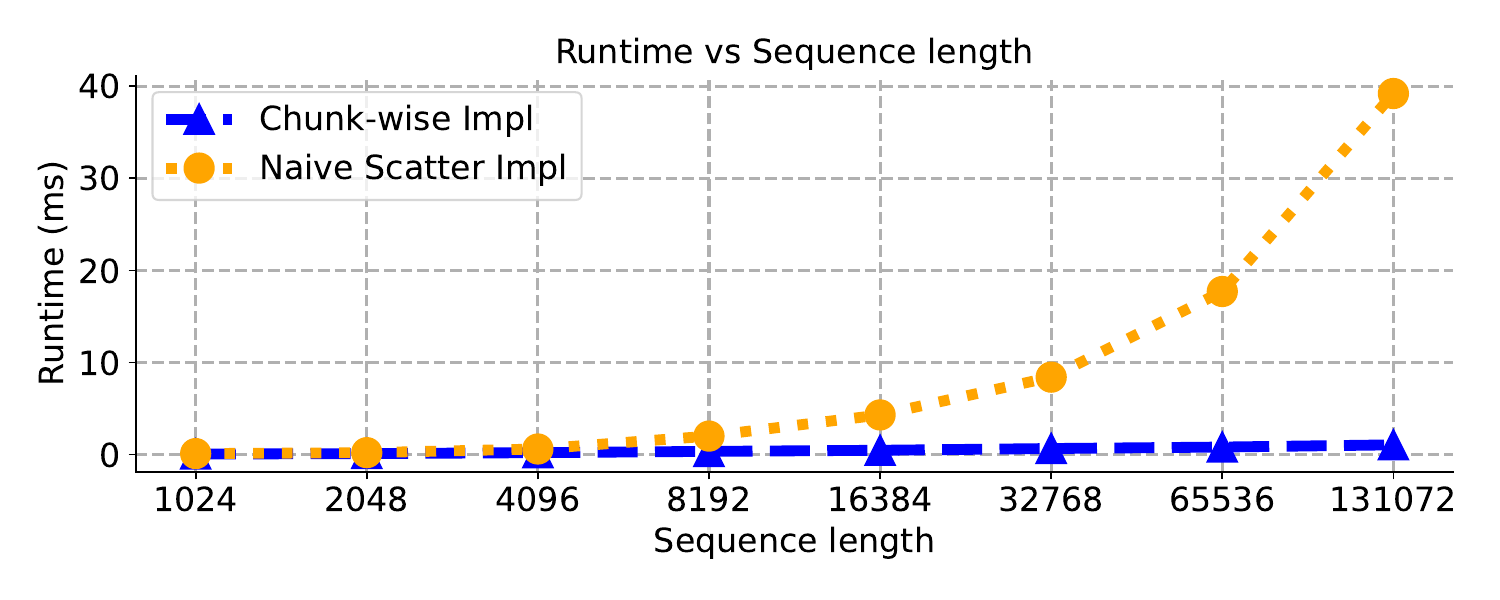}
    \includegraphics[width=0.8\columnwidth]{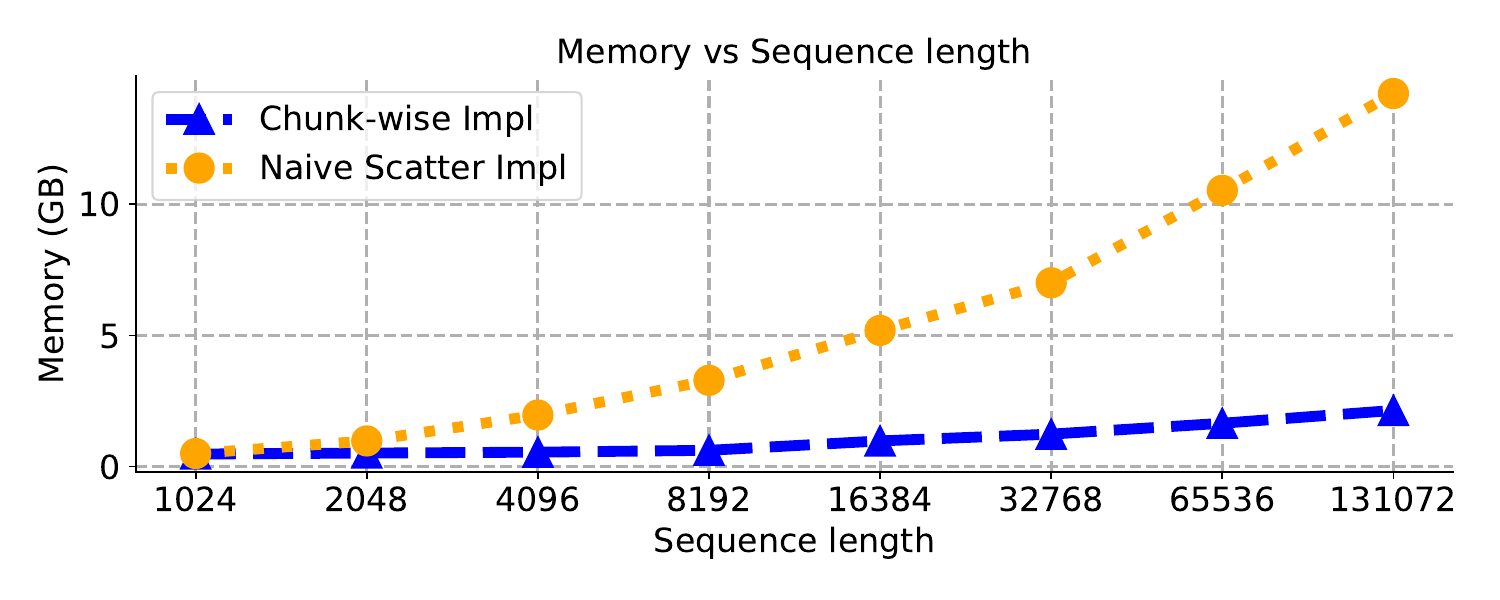}
    \caption{Comparison between between scatter-based implementation (yellow) and our chunk-wise implementation (blue).}
    \vspace{-3mm}
    \label{fig:triton}
\end{figure}

In our implementation, we use Triton \cite{triton} to perform chunk-wise matrix multiplication. As illustrated in Figure \ref{fig:triton}, it outperforms the naive scattered-based operation, demonstrating improved speed and reduced memory usage when computing attention for extremely long sequences. 

\subsection{Cross Window Interaction}
\begin{figure}[t]
    \centering
    \includegraphics[width=.7\columnwidth]{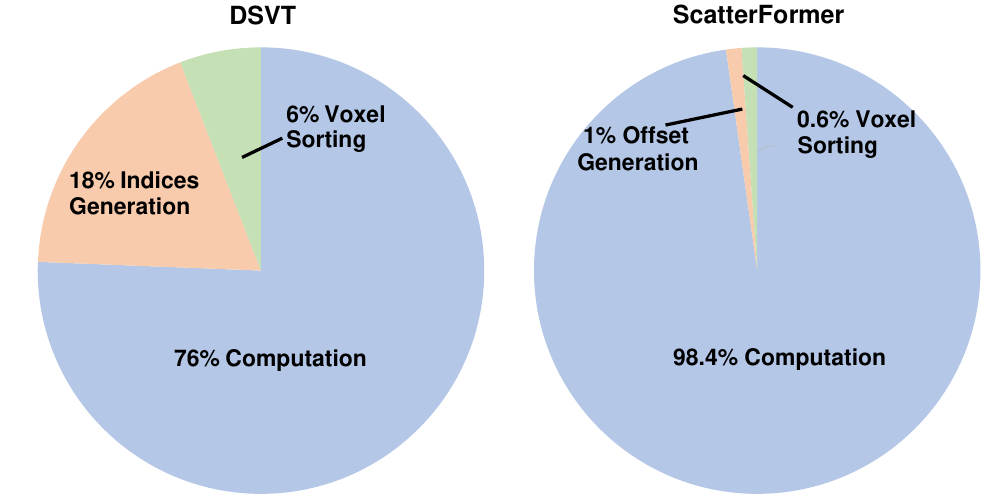}
    
    \caption{Runtime decomposition in DSVT \cite{dsvt} and ScatterFormer backbones. The offsets generation means computing the starting address of each chunk in the flattened matrix.}
    \vspace{-3mm}
    \label{fig:overhead}
\end{figure}

Traditional window-based transformers \cite{sst, dsvt, swformer} utilize window shifting for inter-window connection. For point cloud sequences, this requires recalculating window coordinates and rearranging the voxels. In DSVT, the computation of voxel indices in subsets and the rearrangement of voxel matrices are particularly time consuming, accounting for \textbf{ 24\%} of the total backbone latency, as shown in Figure \ref{fig:overhead}. To alleviate this unnecessary computational overhead from window shifting, we introduce the Cross-Window Interaction (CWI) module, which employs convolutions with large kernel to blend voxel features across windows. 

We adhere to the design principles of the Inception architecture \cite{inception}, where the input features are divided along the channel dimension and processed through multiple branches using group-wise convolutions. Instead of employing a single large kernel for convolution, we adopt more efficient 1D kernels along different axes for feature aggregation. Specifically, we apply a $(S_h+1) \times 1 \times 1$ convolution in one branch and a $1 \times (S_w+1) \times 1$ convolution in another branch, where $(S_w, S_h)$ denotes the window size. Additionally, we incorporate a standard $3 \times 3 \times 3$ convolution in one branch to enhance locality, while another branch performs an identity mapping. Given input $X$ of channel $4c$, the output $X'$ of the Cross Window Interaction module can be written as:
\begin{equation}
    X' = \text{Concat}(X_{k},X_{w},X_{h}, X^{3c:4c}), 
\end{equation}
where
\begin{align}
X_{k} &= \text{DWConv}_{(S_h+1)\times 1\times 1}(X^{:c}), \\
X_{w} &= \text{DWConv}_{1 \times(S_w+1)\times 1}(X^{c:2c}), \\
X_{h} &= \text{DWConv}_{3 \times 3 \times 3}(X^{2c:3c}).
\end{align}
In our experiments, we found that this axis-decomposed large kernel design can achieve a better precision-latency trade-off. With these lengthy kernels, the voxel features can be efficiently blended among different windows.

\subsection{Detection Head and Loss}
To complement existing detection heads, ScatterFormer generates dense BEV feature maps from sparse voxel representations by placing them back to their spatial locations and filling the unoccupied positions with zeros. For the BEV network, we simply follow \cite{hednet} by adopting a cascaded CNN with convolutional blocks at multiple levels of strides. In terms of detection head and loss function, we follow the design identical to that used in DSVT \cite{dsvt}. We train the detection model with heatmap estimation, bounding box regression, and incorporate an IoU loss for confidence calibration. On the nuScenes dataset, the detection head follows the architecture used in Transfusion \cite{transfusion}.

\section{Experiments}
\label{sec:experiments}
\subsection{Datasets and Evaluation Metrics}
\noindent \textbf{Waymo Open Dataset (WOD).} This dataset contains 230,000 annotated samples split into 160,000 for training, 40,000 for validation, and 30,000 for testing. It uses two metrics for 3D object detection: mean average precision (mAP) and mAP weighted by heading accuracy (mAPH), further categorized into Level 1 (L1) for objects detected by more than five LiDAR points and Level 2 (L2) for those detected with at least one point.

 \textbf{NuScenes.} This dataset comprises 40,000 annotated samples, with 28,000 for training, 6,000 for validation, and 6,000 for testing. On this dataset, the model performance is measured by mean average precision (mAP) across multiple distance thresholds (0.5, 1, 2, and 4 meters) and the nuScenes detection score (NDS), which  combines mAP with a weighted sum of five additional metrics assessing true positive predictions in translation, scale, orientation, velocity, and attribute accuracy.
 
\begin{figure}[h]
    \centering
    \includegraphics[width=0.76\columnwidth]{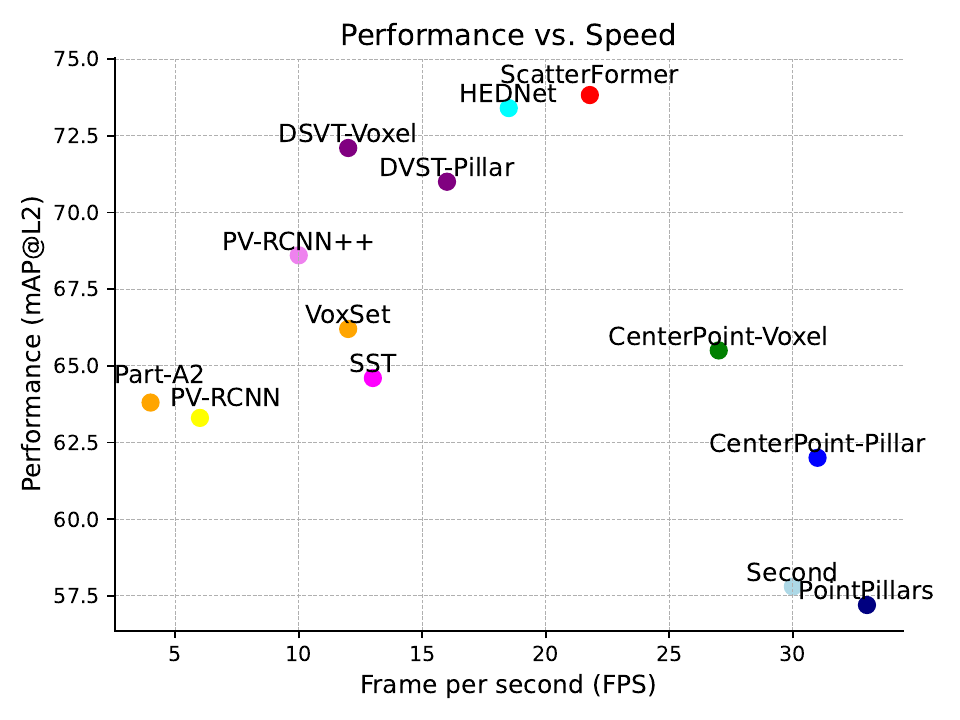}
    \vspace{-2mm}
    \caption{The detection performance (mAPH/L2) vs. speed (FPS) of different methods on Waymo validation set. The speeds are measured on an NVIDIA A100 GPU.}
    \vspace{-5mm}
    \label{fig:tradeoff}
\end{figure}

\subsection{Implementation Details}
Our approach is implemented using the open-source framework OpenPCDet \cite{openpcdet}. To construct ScatterFormer, we set the voxel size to (0.32m, 0.32m, 0.1875m) for the Waymo dataset and (0.3m, 0.3m, 8m) for the NuScenes dataset. The window sizes $(S_w, S_h)$ for the two datasets are set to $(12, 12)$ and $(20, 20)$, respectively. We stack six building blocks for the backbone network. We configure our attention module to have 4 heads with a dimensionality of 128. ScatterFormer is trained for 24 epochs with a learning rate of 0.006 on Waymo Dataset and 20 epochs with a learning rate of 0.004 on NuScenes Dataset. In the last 4 epochs, we disabled the data augmentation strategy of ground-truth sampling. The model is trained on 8 RTX A6000 GPUs with a batch size of 32. Other settings for training and inference adhere strictly to DSVT \cite{dsvt}.

\begin{table*}[t]
		\caption{Performance comparison on the validation set of Waymo Open Dataset. Symbol `*' denotes the methods with temporal modeling, and `-' means that the result is not available.}
		\centering
		\begin{adjustbox}{width=\textwidth}
			\begin{tabular}{c|c|cc|cc|cc|cc}
				\hline \hline
            \rowcolor{gray!20}
            \cellcolor{gray!20} & \cellcolor{gray!20} & \multicolumn{2}{c|}{ALL (3D mAPH)} & \multicolumn{2}{c|}{Vehicle (AP/APH)} & \multicolumn{2}{c|}{Pedestrian (AP/APH)} & \multicolumn{2}{c}{Cyclist (AP/APH)} \\
            \rowcolor{gray!20}
            \multirow{-2}{*}{\cellcolor{gray!20}Method} & \multirow{-2}{*}{\cellcolor{gray!20}Frames} & L1 & L2 & L1 & L2 & L1 & L2 & L1 & L2 \\
				\toprule[1pt]
				
				SECOND \cite{second} & 1 &63.05& 57.23 &72.27/71.69& 63.85/63.33& 68.70/58.18& 60.72/51.31& 60.62/59.28& 58.34/57.05 \\
				PointPillar \cite{pointpillars}& 1& 63.33& 57.53& 71.60/71.00& 63.10/62.50& 70.60/56.70& 62.90/50.20& 64.40/62.30& 61.90/59.90\\
				IA-SSD \cite{ia-ssd} & 1& 64.48&58.08&70.53/69.67& 61.55/60.80 &69.38/58.47& 60.30/50.73& 67.67/65.30 &64.98/62.71\\
				LiDAR R-CNN \cite{lidar-rcnn}& 1& 66.20& 60.10& 73.50/73.00& 64.70/64.20& 71.20/58.70& 63.10/51.70& 68.60/66.90& 66.10/64.40\\
				RSN \cite{rsn}& 1& -& -& 75.10/74.60& 66.00/65.50& 77.80/72.70& 68.30/63.70& -& -\\
				Part-A2 \cite{part-aware}& 1 &70.25& 63.84& 77.05/76.51& 68.47/67.97& 75.24/66.87& 66.18/58.62& 68.60/67.36& 66.13/64.93\\
				Centerpoint \cite{centerpoint} &1& -& 65.50& - & \quad \quad -/66.20& -& \quad \quad -/62.60& -& \quad \quad -/67.60\\
				VoTR \cite{votr} & 1 & -&-&74.95/74.25&65.91/65.29&-&-&-&-\\
				VoxSeT \cite{voxset} & 1 & 72.24 & 66.22 &74.50/74.03&65.99/65.56&80.03/72.42&72.45/65.39&71.56/70.29&68.95/67.73\\
				SST-1f\cite{sst}&1& -& -&76.22/75.79& 68.04/67.64&81.39/74.05& 72.82/65.93&-&-\\
				SWFormer-1f\cite{swformer} &1& -& -&77.8/77.3& 69.2/68.8&80.9/72.7 &72.5/64.9 &-&- \\
				PV-RCNN \cite{pvrcnn}& 1& 69.63& 63.33& 77.51/76.89& 68.98/68.41& 75.01/65.65& 66.04/57.61& 67.81/66.35& 65.39/63.98\\
				PillarNet \cite{pillarnet}& 1& 74.60&68.43&79.09/78.59& 70.92/70.46& 80.59/74.01& 72.28/66.17& 72.29/71.21& 69.72/68.67\\
				PV-RCNN++ \cite{pvrcnn} &1& 75.21& 68.61& 79.10/78.63& 70.34/69.91& 80.62/74.62& 71.86/66.30& 73.49/72.38& 70.70/69.62\\
                FlatFormer \cite{flatformer} & 1 & - & 67.2 &-& 69.0/68.6 &-& 71.5/65.3 &-& 68.6/67.5 \\
                PillarNext-1f \cite{pillarnext} & 1 & 75.74& 69.74&78.40/77.90& 70.27/69.81&82.53/77.14& 74.90/69.80 &73.21/72.20&70.58/69.62 \\
                VoxelNext \cite{voxelnext} & 1 & 76.3 & 70.1 & 78.2/77.7 & 69.9/69.4 &81.5/76.3 &73.5/68.6 &76.1/74.9 &73.3/72.2\\
                FSD \cite{fsd} & 1 & 77.3 & 70.8 & 79.2/78.8 & 70.5/70.1 & 82.6/77.3 & 73.9/69.1 & 77.1/76.0 & 74.4/73.3 \\
   	        DSVT-1f \cite{dsvt} &1& 78.2& 72.1& 79.7/79.3& 71.4/71.0& 83.7/78.9& 76.1/71.5& 77.5/76.5& 74.6/73.7\\
                HEDNet-1f \cite{hednet} & 1 & 79.5 & 73.4 & \textbf{81.1}/\textbf{80.6} & \textbf{73.2}/\textbf{72.7} & 84.4/\textbf{80.0} & 76.8/\textbf{72.6} & 78.7/77.7 & 75.8/74.9 \\

\textbf{ScatterFormer} (ours) & 1 & \textbf{79.7} & \textbf{73.8} & 81.0/80.5 & 73.1/72.7 & \textbf{84.5}/79.9 & \textbf{77.0}/72.6 & \textbf{79.9}/\textbf{78.9} & \textbf{77.1}/\textbf{76.1} \\

				\toprule[1pt]
                SST-3f\cite{sst}&3& -& -&78.66/78.21& 69.98/69.57&83.81/80.14& 75.94/72.37&-&-\\
                FlatFormer-3f \cite{flatformer} & 3 & - & 72.0 &-& 71.4/71.0 &-& 74.5/71.3 &-& 74.7/73.7\\
				SWFormer-3f\cite{swformer}&3& -& -&79.4/78.9& 71.1/70.6&82.9/79.0& 74.8/71.1&-&-\\
                PillarNext-3f \cite{pillarnext} & 3 & 80.0& 74.5& 80.6/80.1 & 72.9/72.4 &85.0/82.1 &78.0/75.2 & 78.9/77.9 & 76.7/75.7\\
                DSVT-4f \cite{dsvt} &4& 81.3& 75.6& 81.8/81.4& 74.1/73.6& 85.6/82.8& 78.6/75.9& 80.4/79.6 & 78.1/77.3\\

                \textbf{ScatterFormer-3f} (ours)  & 3 &81.7 & 76.0 & 82.0/81.4 & 75.0/ 74.1 & 85.7/83.2 & 79.0/76.1 & 80.6/79.5 & 78.1/ 77.7 \\
                
                \textbf{ScatterFormer-4f} (ours)  & 4 & \textbf{81.9} & \textbf{76.7}  &  \textbf{82.4}/\textbf{81.9} & \textbf{75.2}/\textbf{74.7} & \textbf{86.1}/\textbf{83.4} & \textbf{79.3}/\textbf{76.6} & \textbf{81.2}/\textbf{80.4} & \textbf{78.9}/\textbf{78.1} \\

                \hline\hline
                *3D-MAN \cite{3dman} &16& -& -& 74.53/74.03& 67.61/67.14& 71.7/67.7& 62.6/59.0&-&-\\
                *CenterFormer \cite{centerformer} & 4 & 77.0 & 73.2 &78.1/77.6& 73.4/72.9& 81.7/78.6& 77.2/74.2& 75.6/74.8& 73.4/72.6\\  
				*CenterFormer \cite{centerformer} & 8 & 77.3 & 73.7 &78.8/78.3& 74.3/73.8& 82.1/79.3& 77.8/75.0& 75.2/74.4& 73.2/72.3 \\ 
				*MPPNet \cite{mppnet} &4& 79.83& 74.22& 81.54/81.06& 74.07/73.61& 84.56/81.94& 77.20/74.67& 77.15/76.50& 75.01/74.38\\
				*MPPNet \cite{mppnet} &16& 80.40& 74.85& 82.74/82.28& 75.41/74.96& 84.69/82.25& 77.43/75.06& 77.28/76.66& 75.13/74.52\\
				*MSF\cite{msf} & 4 & 80.20 & 74.62 & 81.36/80.87& 73.81/73.35& 85.05/82.10& 77.92/75.11& 78.40/77.61& 76.17/75.40\\
				*MSF\cite{msf} & 8 & 80.65 & 75.46& 82.83/82.01& 75.76/75.31&  85.24/82.21& 78.32/75.61&  78.52/77.74& 76.32/75.47\\
                \toprule[1pt]
			\end{tabular}
		\end{adjustbox}
		\label{tbl:val}
		\vspace{-7mm}
\end{table*}

\begin{table*}[t]
		\caption{Performance comparison on the test set of Waymo Open Dataset. `-' means that the result is not available.}
  \vspace{-2mm}
		\centering
		\begin{adjustbox}{width=\textwidth}
			\begin{tabular}{c|cc|cc|cc|cc}
				\hline \hline
            \rowcolor{gray!20}
            \cellcolor{gray!20} & \multicolumn{2}{c|}{ALL (3D mAPH)} & \multicolumn{2}{c|}{Vehicle (AP/APH)} & \multicolumn{2}{c|}{Pedestrian (AP/APH)} & \multicolumn{2}{c}{Cyclist (AP/APH)} \\
            \rowcolor{gray!20}
            \multirow{-2}{*}{\cellcolor{gray!20}Method} & L1 & L2 & L1 & L2 & L1 & L2 & L1 & L2 \\
            \hline
				PointPillar\cite{pointpillars}&-&-&68.10& 60.10& 68.00/55.50& 61.40/50.10&-&-\\
				StarNet\cite{starnet}&-&-&61.00& 54.50& 67.80/59.90& 61.10/54.00 &-&-\\
				M3DETR\cite{m3detr} & 67.1 &61.9 &77.7/77.1& 70.5/70.0& 68.2/58.5& 60.6/52.0& 67.3/65.7& 65.3/63.8\\ 
				3D-MAN \cite{3dman} & -& -& 78.28& 69.98& 69.97/65.98& 63.98/60.26& -& -\\
				PV-RCNN++ \cite{pvrcnn++} & 75.7& 70.2&81.6/81.2& 73.9/73.5& 80.4/75.0& 74.1/69.0& 71.9/70.8& 69.3/68.2 \\
				CenterPoint \cite{centerpoint}& 77.2& 71.9& 81.1/80.6 &73.4/73.0& 80.5/77.3& 74.6/71.5& 74.6/73.7& 72.2/71.3 \\
				RSN \cite{rsn} & -& -& 80.30& 71.60& 78.90/75.60 &70.70/67.80 &-&-\\
				SST-3f \cite{sst} &78.3 &72.8 &81.0/80.6 &73.1/72.7& 83.3/79.7 &76.9/73.5 &75.7/74.6 &73.2/72.2 \\
                HEDNet \cite{hednet} & 79.05 & 73.77 &83.78/83.39&76.33/75.96 &83.46/78.98	& 77.53/73.25&75.86/74.78 & 73.13/72.09\\
                PillarNext-3f \cite{pillarnext} &79.00 & 74.09 & 83.28/82.83 & 76.18/75.76 & 84.40/81.44	& 78.84/75.98 & 73.77/72.73	& 71.56/70.55\\
				\toprule[1pt]
			
\textbf{ScatterFormer-4f} (ours) &\textbf{80.71}&\textbf{75.84}& \textbf{85.60}/\textbf{85.13} &\textbf{78.34}/\textbf{78.07} & \textbf{84.58}/\textbf{81.64} &\textbf{79.12}/\textbf{76.32}&\textbf{76.20}/\textbf{75.35}& \textbf{74.02}/\textbf{73.13}\\

				\toprule[1pt]
			\end{tabular}
		\end{adjustbox}
		\label{tbl:test}
		
\end{table*}

\begin{table*}
\centering
\caption{The performance on the validation set of NuScenes.}
 \vspace{-2mm}
\begin{adjustbox}{width=\textwidth}
\begin{tabular}{c|c|c|c|c|c|c|c|c|c|c|c|c}
\hline	\hline\rowcolor{gray!20}
Method & NDS & mAP & Car & Truck & Bus & T.L. & C.V. & Ped. & M.T. & Bike & T.C. & B.R. \\ 
\hline
CenterPoint\cite{centerpoint} & 66.5 & 59.2 & 84.9 & 57.4 & 70.7 & 38.1 & 16.9 & 85.1 & 59.0 & 42.0 & 69.8 & 68.3 \\
VoxelNeXt\cite{voxelnext} & 66.7 & 60.5 & 83.9 & 55.5 & 70.5 & 38.1 & 21.1 & 84.6 & 62.8 & 50.0 & 69.4 & 69.4 \\
TransFusion-L\cite{transfusion} & 70.1 & 65.5 & 86.9 & 60.8 & 73.1 & 43.4 & 25.2 & 87.5 & 72.9 & 57.3 & 77.2 & 70.3 \\
PillarNext\cite{pillarnext} & 68.4 & 62.2 & 85.0 & 57.4 & 67.6 & 35.6 & 20.6 & 86.8 & 68.6 & 53.1 & 77.3 & 69.7 \\
DSVT\cite{dsvt} & 71.1 & 66.4 & 87.4 & 62.6 & 75.9 & 42.1 & 25.3 & 88.2 & 74.8 & 58.7 & 77.8 & 70.9 \\
HEDNet \cite{hednet} &  71.4 & 66.7 & 87.7 & 60.6 &77.8 &\textbf{50.7} &28.9 &87.1 &74.3 &56.8 &76.3 &66.9\\
\hline
\textbf{ScatterFormer} (ours) & \textbf{72.4} & \textbf{68.3} & \textbf{88.6} & \textbf{65.4} & \textbf{79.3} & 45.4 & \textbf{29.1} & \textbf{88.7} & \textbf{74.7} & \textbf{61.5} & \textbf{78.2} & \textbf{72.3} \\
 	\hline
\end{tabular}
\end{adjustbox}
\label{tbl:nuscenes}
\vspace{-5mm}
\end{table*}

\begin{table}[h]
		\centering
		\caption{Ablation experiments on the validation set of Waymo Open Dataset. ``SLA'' and ``CWI'' refer to the proposed Scattered Linear Attention and Cross-Window Interaction modules, respectively. ``CPE" refers to Conditional Position Embedding. AP and APH scores on LEVEL2 are reported. }
  \vspace{-2mm}
		\begin{adjustbox}{width=0.65\columnwidth}
		\begin{tabular}{c|c|ccc}
		\hline	\rowcolor{gray!20}
	    & Ablation & Veh. & Ped. & Cyc.   \\ \hline\hline
          - & baseline & 71.6/71.1 & 76.0/70.9 & 74.2/73.1\\
        (a) & w/o SLA module & 68.5/68.0 & 73.3/67.0 & 71.0/69.9\\
        (b) & w/o CWI module & 69.6/69.2 & 74.0/69.9 & 72.1/71.1\\
        (c) & CWI $\rightarrow$ SW & 69.9/69.5 & 75.0/70.4 & 73.8/72.4\\
        (d) & w/o CPE & 70.2/69.8 & 73.3/67.2 & 72.4/70.5\\
		\hline
		\end{tabular}
		\end{adjustbox}
		\label{tbl:abl}
\end{table}

 \begin{table}[h]
\centering
\caption{Comparison of model performance using different window sizes at various stages. The APH (L2) scores on different categories are reported. }
  \vspace{-2mm}
\begin{adjustbox}{width=.52\columnwidth}
\begin{tabular}{c|c|ccc}
\hline \rowcolor{gray!20}
Window Size     & Region Size & Veh. & Ped. & Cyc. \\
\hline\hline
10 & 3.20 m & 70.2 & 69.2 & 71.0 \\
12 & 3.84 m  & 71.1   & 70.9 & 73.1 \\
14 & 4.48 m   & 71.3   & 69.9 & 72.5 \\
16 & 5.12 m   & 71.0   & 69.2 & 72.3 \\

\hline
\end{tabular}
\end{adjustbox}
\label{tbl:window}
\vspace{-2mm}
\end{table}

\begin{table}[h]
\centering
\caption{Performance with various kernel configurations in the Cross-Window Integration (CWI) module. Each configuration is specified by the notation ``KxDy", where ``Kx" refers to the kernel size, and ``Dy" denotes the dilation rate. The APH (L2) scores on different categories are reported.}
  \vspace{-2mm}
\begin{adjustbox}{width=.5\columnwidth}
\begin{tabular}{l|c|ccc}
\hline \rowcolor{gray!20}
Kernel & Latency & Veh. & Ped. & Cyc. \\
\hline\hline
$\text{Conv1D}_{\text{K13}} \times 2$ & 47ms & 71.1   & 70.9 & 73.1 \\
$\text{Conv2D}_{\text{K5D1}}$ & 49ms & 69.5 & 69.0 & 71.1 \\
$\text{Conv2D}_{\text{K5D2}}$ & 49ms & 70.7 & 70.4 & 72.8 \\
$\text{Conv2D}_{\text{K7D1}}$ & 65ms & 71.5 & 71.0 & 72.5 \\

\hline
\end{tabular}
\end{adjustbox}
\label{table:kernel}
\vspace{-2mm}
\end{table}

\subsection{Comparison with State-of-the-Arts}
\noindent \textbf{Results on Waymo Open Dataset (WOD).} We conduct a detailed comparison of ScatterFormer against the published results on the validation set of the Waymo Open Dataset. In line with standard practices, we distinctly categorize and list the methods utilizing single-frame and multi-frame approaches. To ensure comprehensive coverage, we also include comparisons with methods incorporating long-term temporal modeling, such as those described in \cite{centerformer, mppnet, msf}. As demonstrated in Table \ref{tbl:val}, our single-stage model outperforms most two-stage methods and achieves better performance than state-of-the-art methods such as DSVT \cite{dsvt} and HEDNet \cite{hednet}. Moreover,our model achieves 76.0 and 76.7 level 2 mAPH on 3 and 4 frame settings, respectively, outperforming previous multiframe methods by a margin of 1.1. Furthermore, as shown in Figure \ref{fig:tradeoff}, our method significantly enhances the detection runtime. This improvement is attributed to our use of Linear Attention, which eliminates the need for extensive voxel sorting and padding operations.
Furthermore, as presented in Table \ref{tbl:test}, ScatterFormer achieves the highest Level 1 and Level 2 mAPH scores in all categories of the Waymo test set, highlighting its superior detection quality and precision. Remarkably, our approach, as a one-shot method, delivers performance that is on par with methods that require temporal modeling. This finding underscores the critical importance of network design and architecture in achieving high-performance architecture. 

 \textbf{Results on NuScenes.} We compare ScatterFormer with the previous best performing methods on the nuScenes dataset. As shown in Table \ref{tbl:nuscenes}, ScatterFormer achieves the state-of-the-art performance in terms of \textit{val} NDS (72.4) and mAP (68.3), surpassing HEDNet \cite{hednet} by 1.0 and 1.6 respectively. Compared to TransFusion \cite{transfusion}, ScatterFormer achieves a significant improvement of 2.8\% in mAP, demonstrating that its backbone network is superior to the sparse convolutional network. ScatterFormer exhibits particularly strong performance on categories like Bus, Bike, Construction Vehicle and Trailer, suggesting its ability to effectively capture and encode detailed contextual features.

\begin{table}[h]
\centering
\caption{Performance with different linear attention designs. The APH (L2) scores on different categories are reported.}
  \vspace{-2mm}
\begin{adjustbox}{width=.5\columnwidth}
\begin{tabular}{l|ccc}
\hline \rowcolor{gray!20}
Linear Attention &  Veh. & Ped. & Cyc. \\
\hline\hline
Efficient Attn \cite{efficientvit} & 71.1   & 70.9 & 73.1  \\
Gated Linear Attn \cite{transnormerv2} & 69.9 & 69.0 & 72.5 \\
Focused Linear Attn \cite{flattenTr} &  70.4 & 69.5 & 72.3 \\ 
XCA \cite{xcit} & 71.5   & 70.4  & 72.8 \\
\hline
\end{tabular}
\end{adjustbox}
\label{tbl:attn}
\vspace{-4mm}
\end{table}

\subsection{Ablation Study}

In this section, we conduct ablation experiments in ScatterFormer using 20\% of the Waymo training and validation data. Each model was trained for 24 epochs and run on default parameters. Table \ref{tbl:abl} shows four different configurations of ScatterFormer: (a) removing the Scattered Linear Attention module (SLA), (b) removing the Cross Window Interaction (CWI) module, (c) replacing the CWI with Shifted Window (SW) approach, and (d) removing Conditional Position Encoding (CPE) module.

As can be seen in Table \ref{tbl:abl}, configuration (a) results in ScatterFormer experiencing a decrease of (3.1\%, 3.9\%, and 3.2\%) APH on three categories.This highlights the effectiveness of the SLA, as it successfully captures long-range contexts and facilitates dynamic feature extraction for object detection. In configuration (b), ScatterFormer, which lacks cross-window interaction capability, exhibits a performance drop of (1.9\%, 1.0\%, and 2.0\%) APH across the three categories. While the Shifted Window approach in configuration (c) enhances performance, the improvement is not as substantial as that achieved by the proposed CWI module. The results in configuration (d) indicate that CPE contributes to a certain degree of performance enhancement.

\textbf{Window Size.}
In Table \ref{tbl:window}, We tested the impact of different window sizes on the detection performance of various target types. It can be observed that ScatterFormer is not highly sensitive to window size. Slightly increasing the window size can enhance the detection performance for small targets, indicating that for objects like pedestrians and bicycles, where the point cloud is overly sparse, appropriate contextual information can enhance the recognition of these objects. However, excessively large windows can degrade the performance. We speculate that this is because larger windows encompass more tokens, thereby mitigating the focusing power of linear attention.

\vspace{2mm}

\textbf{Kernel Design in CWI.}
Table \ref{table:kernel} presents the performance under various kernel configurations in the Cross-Window Interaction (CWI) module. Increasing the number of parameters in the kernel significantly enhances performance. For example, using a large 7$\times$7 kernel brings notable performance improvements, but also introduces more latency. Additionally, larger kernel strides enhance the exchange rate of tokens across windows. For example, using dilated convolutions in CWI performs slightly better than using non-dilated convolutions. Instead of scaling up large, we decompose large kernel into more efficient 1D kernels, which achieves the optimal latency while keeping comparable performance. As the current Spconv library lacks support for depth-wise convolution, we have developed a modified version to address this limitation.\footnote{\url{https://github.com/skyhehe123/spconv}}

\subsection{Comparison with Different Linear Attentions}
We further explored the performance of ScatterFormer by conducting experiments with several other linear attention mechanisms, including Gated Linear Attention \cite{transnormerv2}, Cross-covariance Attention (XCA) \cite{xcit}, and Focused Linear Attention \cite{flattenTr}. The results are summarized in Table \ref{tbl:attn}. Our findings suggest that the window-based attention framework employed by ScatterFormer is not highly sensitive to the specific form of linear attention used. Notably, Efficient Attention \cite{efficientvit} performs slightly worse than XCA \cite{xcit} in the Vehicle class but better in the Pedestrian and Cyclist classes. Focused Linear Attention introduces additional computational overhead, while Gated Linear Attention adds more learnable parameters to the model.
\vspace{-2mm}
\subsection{Limitations}
ScatterFormer relies on our customized operators, which have not yet been implemented as plugins in TensorRT. Therefore, deploying ScatterFormer on in-vehicle devices will require additional engineering efforts. Despite this, ScatterFormer is more efficient than current models based on sparse convolution and traditional attention when used on consumer-grade GPU cards. Furthermore, ScatterFormer can be optimized by dynamically partitioning matrices according to different GPU architectures to leverage TensorCore for hardware acceleration.

\section{Conclusion}
\label{sec:conclusion}
We introduced ScatterFormer, an innovative architecture designed for 3D object detection using point clouds, specifically targeting the challenges associated with processing sparse and unevenly distributed data from LiDAR sensors. The cornerstone of our approach is the Scattered Linear Attention (SLA) module, which effectively addresses the limitations of conventional attention mechanisms in managing voxel features of varying lengths. SLA ingeniously combines linear attention with a chunk-wise matrix multiplication algorithm, tailored to meet the distinct requirements of processing voxels grouped by windows. By integrating SLA with a novel cross-window interaction (CWI) module, ScatterFormer achieves higher accuracy and lower latency, surpassing traditional transformer-based and sparse CNN-based detectors in extensive 3D detection tasks.

% \clearpage\mbox{}Page \thepage\ of the manuscript.
% \clearpage\mbox{}Page \thepage\ of the manuscript.
% \clearpage\mbox{}Page \thepage\ of the manuscript.
% \clearpage\mbox{}Page \thepage\ of the manuscript.
% \clearpage\mbox{}Page \thepage\ of the manuscript. This is the last page.
% Now we have reached the maximum length of an ECCV \ECCVyear{} submission (excluding references).
% References should start immediately after the main text, but can continue past p.\ 14 if needed.
% \clearpage  % TODO REVIEW/FINAL: This \clearpage needs to be removed from both review and camera-ready versions.

\bibliographystyle{splncs04}
\bibliography{main}
\end{document}